\documentclass[10pt,twocolumn,letterpaper]{article}

 \usepackage{cvpr}              

\newcommand{\mysection}[1]{\vspace{2pt}\noindent\textbf{#1}}

\newlength\mytmplen
\usepackage{tikz}          
\usepackage{caption}       
\usetikzlibrary{spy}       

\DeclareRobustCommand{\zoomin}[9]{ %
\begin{tikzpicture}[spy using outlines={rectangle,#9,magnification=#8,size=#6}]   
	\node[anchor=south west,inner sep=0]  {\includegraphics[width=#7]{#1}};
	\spy on (#2, #3) in node at (#4,#5);
\end{tikzpicture}
}

\usepackage{cuted}   
\usepackage{graphicx}      
\usepackage{booktabs}      
\usepackage[table]{xcolor} 
\usepackage{colortbl}      
\usepackage{mwe} 

\newcommand{\best}{\cellcolor{tablered}}

\definecolor{tablered}{rgb}{1, 0.7, 0.7}

\providecommand{\methodname}{Triangle Splatting+ }

\usepackage[table]{xcolor} 
\usepackage{pifont}        

\newcommand{\cmark}{\ding{51}}  
\newcommand{\xmark}{\ding{55}}  

\usepackage{subcaption} 
\definecolor{cvprblue}{rgb}{0.21,0.49,0.74}
\usepackage[pagebackref,breaklinks,colorlinks,allcolors=cvprblue]{hyperref}

\def\paperID{*****} 
\def\confName{CVPR}
\def\confYear{2025}

\title{Triangle Splatting+: Differentiable Rendering with Opaque Triangles}

\author{
\textbf{Jan Held}$^{1,2}$\quad
\textbf{Renaud Vandeghen}$^{1}$\quad
\textbf{Sanghyun Son}$^{3}$\quad \\
\textbf{Daniel Rebain}$^{4}$\quad 
\textbf{Matheus Gadelha}$^{6}$\quad
\textbf{Yi Zhou}$^{6}$\quad \\
\textbf{Ming C. Lin}$^{3}$\quad
\textbf{Marc Van Droogenbroeck}$^{1}$\quad
\textbf{Andrea Tagliasacchi}$^{2,5}$ \\[0.5em]
$^1$University of Liège \quad
$^2$Simon Fraser University \quad
$^3$University of Maryland \quad \\
$^4$University of British Columbia \quad
$^5$University of Toronto \quad
$^6$Adobe Research \\
}

\begin{document}
\maketitle
\begin{strip}
\centering
\includegraphics[width=\textwidth]{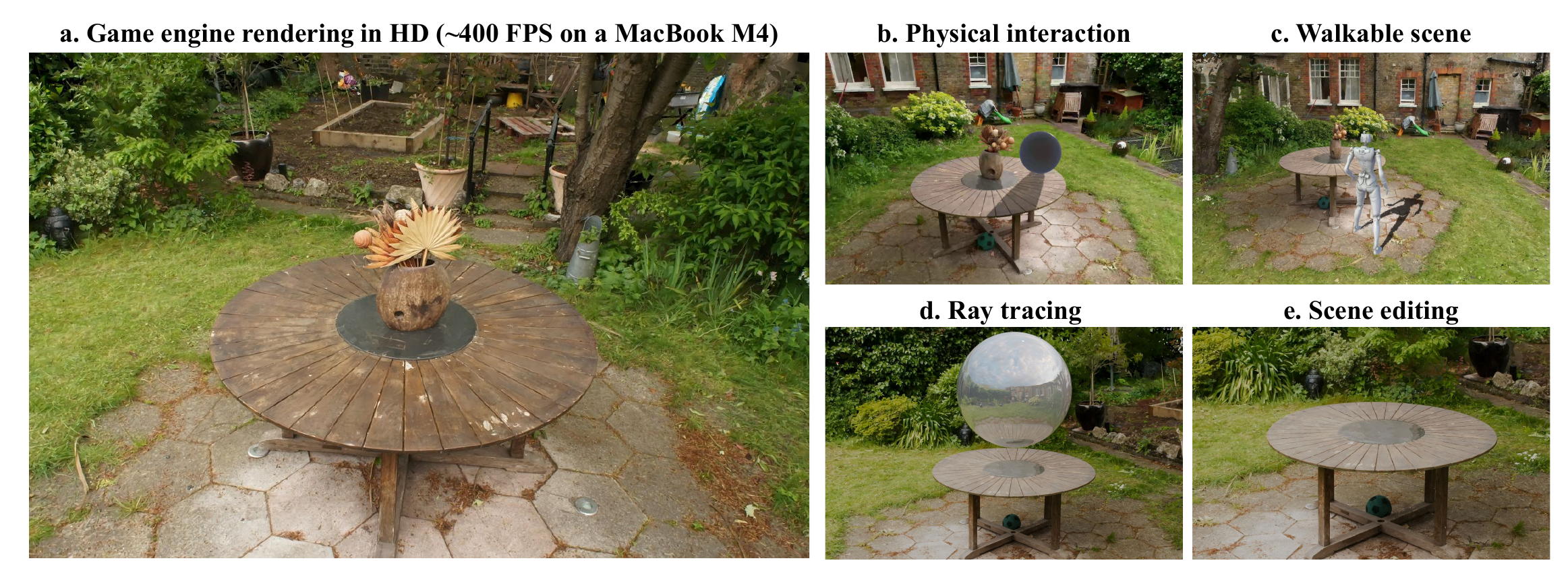}
\captionof{figure}{\small
\textbf{\methodname}optimizes a triangle-based representation end-to-end and achieves high visual quality using only \textit{opaque triangles.} The resulting representation can be directly imported into any game engine without post-processing and runs at 400 FPS on a consumer laptop. Within the engine, it naturally supports (b) physical interactions (e.g., collisions), (c) walkable scene interactions as in video games, (d) ray tracing, and (e) scene editing (e.g., object removal or addition).
}
\label{fig:teaser}
\end{strip}
\begin{abstract}

Reconstructing 3D scenes and synthesizing novel views has seen rapid progress in recent years. Neural Radiance Fields demonstrated that continuous volumetric radiance fields can achieve high-quality image synthesis, but their long training and rendering times limit practicality. 3D Gaussian Splatting (3DGS) addressed these issues by representing scenes with millions of Gaussians, enabling real-time rendering and fast optimization.  
However, Gaussian primitives are not natively compatible with the mesh-based pipelines used in VR headsets, and real-time graphics applications. Existing solutions attempt to convert Gaussians into meshes through post-processing or two-stage pipelines, which increases complexity and degrades visual quality.  
In this work, we introduce \textbf{\methodname}, which directly optimizes triangles, the fundamental primitive of computer graphics, within a differentiable splatting framework. We formulate triangle parametrization to enable connectivity through shared vertices, and we design a training strategy that enforces opaque triangles. The final output is immediately usable in standard graphics engines without post-processing.
Experiments on the Mip-NeRF360 and Tanks \& Temples datasets show that \methodname achieves state-of-the-art performance in mesh-based novel view synthesis. Our method surpasses prior splatting approaches in visual fidelity while remaining efficient and fast to training. Moreover, the resulting semi-connected meshes support downstream applications such as physics-based simulation or interactive walkthroughs. 
The project page is \href{https://trianglesplatting2.github.io/trianglesplatting2/}{https://trianglesplatting2.github.io/trianglesplatting2/}.

\end{abstract}
    
\section{Introduction}
\label{sec:intro}

In recent years, significant progress has been made in reconstructing complex scenes and generating novel views.
First, Neural Radiance Fields~\cite{Mildenhall2020NeRF-eccv} revolutionized this area by representing scenes as continuous volumetric radiance fields, which are optimized to render novel views at high-quality. 
Despite their impact, NeRF-based methods are hindered by long training times and slow inference, which restrict their practical applicability. To overcome these limitations, 3D Gaussian Splatting~\cite{Kerbl20233DGaussian} was introduced, representing scenes with millions of 3D Gaussian primitives. This representation enables orders-of-magnitude faster training and supports real-time rendering, all while preserving high visual fidelity.

However, Gaussian primitives are not natively compatible with the graphics pipelines that power interactive applications such as VR headsets or real-time mesh-based renderers.
Integrating them requires either modifying existing engines to directly support Gaussians, or devising conversion techniques that transform Gaussian radiance fields into mesh representations.
Several recent works have pursued this direction: 2DGS~\cite{Huang20242DGaussian} and RaDe-GS~\cite{Zhang2024RaDeGS-arxiv} reconstruct meshes by first optimizing a scene and then extracting surfaces via truncated signed distance fields (TSDF), or MiLo~\cite{Guedon2025MILo-arxiv}, which learns a surface mesh jointly during training.
Although effective to some extent, all of these approaches depend on additional post-processing, which complicates the pipeline and often leads to degraded quality.

Held \etal proposed an initial step toward bridging classical rendering pipelines with differentiable radiance field frameworks by replacing the Gaussian primitive by \textit{triangles.}
Triangle Splatting~\cite{Held2025Triangle-arxiv} demonstrated that unstructured sets of triangles can be optimized end-to-end within a differentiable splatting framework, combining the adaptability of Gaussian splats with the efficiency and compatibility of triangles.
Yet, Triangle Splatting represents only an initial step toward bridging classical rendering pipelines with differentiable radiance field frameworks.
When its triangle soup is rendered in a game engine, a noticeable drop in visual quality occurs because training does not enforce opaque triangles but instead relies on soft, semi-transparent ones.
Moreover, all triangles remain isolated: due to the chosen parametrization, no connectivity can be established between neighboring triangles, even when their vertices coincide spatially or share similar color distributions. 

In this work, we introduce \textbf{\methodname}, which unifies radiance field optimization with traditional computer graphics.
We reformulate the parametrization of triangles to enable connectivity: starting from a shared set of vertices, triangles can naturally connect through common vertices rather than remaining isolated.
In addition, we design a training strategy that enforces opaque triangles, ensuring that the final representation integrates seamlessly into mesh-based rendering pipelines.
Although the resulting mesh is only semi-connected, it is sufficient for a wide range of downstream applications, including physics-based simulation, or interactive walkthroughs.
\methodname thus paves the way for immediate integration into VR/AR environments and modern video games. 

\paragraph{Contributions.}

\begin{enumerate*}[label=\textbf{(\roman*)}]
\item We introduce \textbf{\methodname}, which unifies radiance field optimization with traditional computer graphics by obtaining high-visual quality in mesh-based renderer.
\item We re-parametrize triangles to enforce connectivity between primitives
\item We propose a tailored training strategy that balances visual fidelity and geometric accuracy with solid and opaque triangles.
\item \methodname can run on any game-engine and enables physcical simulations, walkable evironments for video games and many more.

\end{enumerate*}

\section{Related work}
\label{sec:related_work}

\begin{figure*}[t]
\centering
\includegraphics[width=0.99\linewidth]{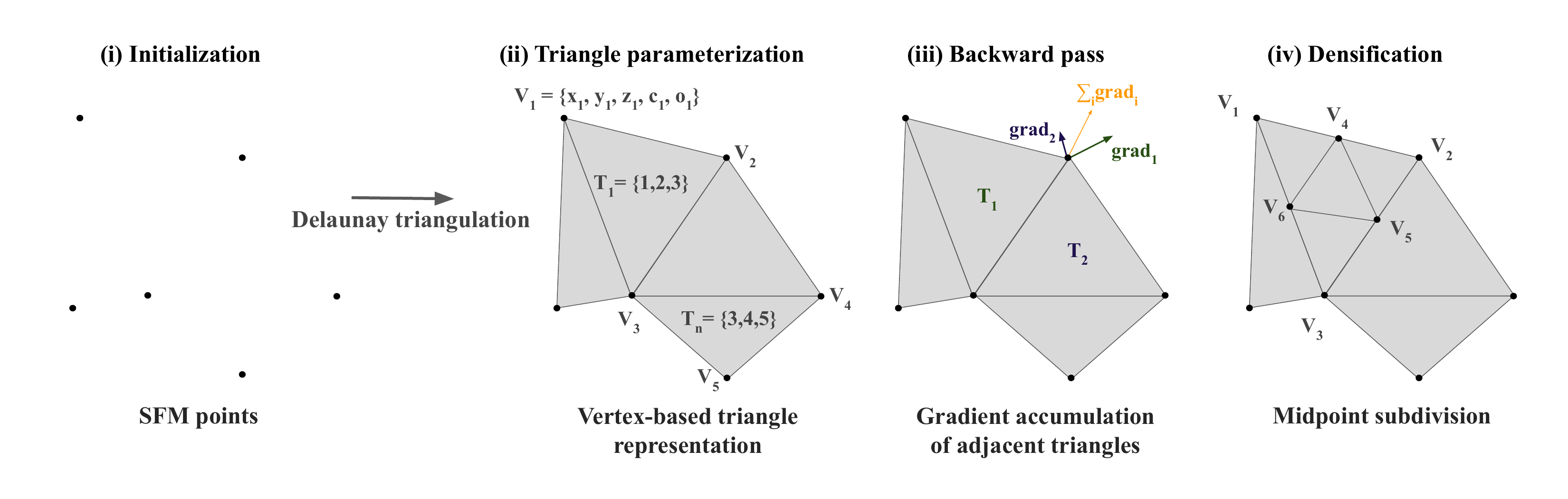}
\caption{\textbf{Method overview of \methodname.} (i) We start from sparse SfM points and apply 3D Delaunay triangulation to obtain an initial mesh.
(ii) Triangles are parameterized through a shared vertex set, where each vertex stores position $x_i, y_i, z_i$, color $c_i$, and opacity $o_i$. Each triangle is defined by the three indices in the vertex set that compose it.
(iii) During training, gradients from all adjacent triangles accumulate at shared vertices.
(iv) Densification is performed by midpoint subdivision, introducing new vertices and triangles while preserving connectivity.}
\label{fig:methodology}
\end{figure*}

\paragraph{Primitive-based differentiable rendering.}
Differentiable rendering enables end-to-end optimization by propagating image-based losses back to scene parameters, allowing for the learning of explicit representations such as point clouds~\cite{Gross2007Point,Kato2018Neural}, voxel grids~\cite{FridovichKeil2022Plenoxels}, polygonal meshes~\cite{Kato2018Neural, Liu2019SoftRasterizer, Loper2014OpenDR}, and more recently, Gaussian primitives~\cite{Kerbl20233DGaussian}.
The advent of 3D Gaussian Splatting~\cite{Kerbl20233DGaussian} showed that it is possible to fit millions of anisotropic Gaussians in just minutes, enabling real-time rendering with high fidelity.
Since then, various directions have been explored to improve the Gaussian primitive, including the use of 2D Gaussians~\cite{Huang20242DGaussian}, generalized Gaussians~\cite{Hamdi2024GES}, alternative kernels~\cite{Huang2025Deformable}, and learnable basis functions~\cite{Chen2024Beyond-arxiv}.
Other works moved beyond Gaussians entirely, investigating different primitives such as smooth 3D convexes~\cite{Held20253DConvex}, linear primitives~\cite{vonLutzow2025LinPrim-arxiv}, sparse voxel fields~\cite{Sun2025Sparse}, or radiance foams~\cite{Govindarajan2025Radiant-arxiv}.
More recently, Held \etal advocated for the comeback of triangles, the most fundamental primitive in computer graphics. 
Many researchers have explored this direction, proposing triangle-based representations for efficient scene modeling~\cite{Burgdorfer2025Radiant-arxiv, Jiang2025HaloGS-arxiv}.
However, existing triangle-based methods usually result in an unstructured triangle soup, with no connectivity between adjacent triangles, and they fail to produce solid, opaque triangles at the end of training.
Another line of work using differentiable mesh for 3D reconstruction~\cite{Son2024DMesh, Son2024DMesh++-arxiv} produce connected, solid mesh, but they mainly handle synthetic object and do not extend to real-world scenes.
In this work, we propose a method that enables triangle connectivity, producing a semi-structured mesh by the end of training. Moreover, our representation consists only of opaque triangles, making the output immediately compatible with game engines.

\paragraph{Mesh reconstruction from images.}

Implicit and explicit methods have made significant progress in reconstructing 3D scenes, but they remain largely incompatible with traditional mesh-based renderers and game engines. Some methods propose strategies to convert implicit radiance fields into meshes. BakedSDF~\cite{Yariv2023BakedSDF} learns a neural SDF and appearance then bakes them into a textured triangle mesh, Binary Opacity Fields~\cite{Reiser2024Binary} drives densities toward near binary opacities so surfaces can be extracted as a mesh and rendered efficiently, and MobileNeRF~\cite{Chen2023MobileNeRF} distills a NeRF into a compact set of textured polygons for real-time mobile rendering.
However, these methods introduce overhead and increase the overall training time.
More recently, several methods have built on Gaussian Splatting and proposed ways to extract a mesh from the optimized Gaussian scene. 2DGS~\cite{Huang20242DGaussian} and RaDe-GS~\cite{Zhang2024RaDeGS-arxiv} rely on Truncated Signed Distance Fields (TSDF) for mesh extraction. Other approaches extract meshes by sampling a surface-aligned Gaussian level set followed by Poisson reconstruction~\cite{Guedon2024SuGaR}, or by defining a Gaussian opacity level set and applying Marching Tetrahedra on Gaussian-induced tetrahedral grids~\cite{Yu2024Gaussian}. All of these methods, however, treat mesh extraction as a separate post-processing step. 
More recently, MiLo~\cite{Guedon2025MILo-arxiv} integrates surface mesh extraction directly into the optimization, jointly refining both the mesh and the Gaussian representation. However, while MiLo optimizes the mesh geometry during training, color still has to be learned separately, for instance through a neural color field applied in a post-processing step. In contrast, \methodname directly optimizes opaque triangles together with their vertex colors, making the result immediately compatible with any mesh-based renderer without requiring additional post-processing.

\section{Methodology}
\label{sec:methodology}

Kerbl \etal \cite{Kerbl20233DGaussian} introduced 3D Gaussian Splatting, representing scenes as large sets of Gaussian primitives that are directly optimized  from multi-view observations. 
Building on this idea, subsequent works have explored alternative primitives: 3D Convex Splatting represents scenes using convex shapes~\cite{Held20253DConvex}, 2DGS employs flat 2D Gaussians~\cite{Huang20242DGaussian} and most recently, Held \etal proposed representing a scene using \textit{triangles}. 
While Triangle Splatting demonstrated that unstructured sets of triangles can be optimized end-to-end, it left an open question of whether triangles can also serve as a practical and high-quality representation for mesh-based rendering. 

In this work, we show that triangles can indeed be optimized end-to-end as opaque primitives, yielding a representation that integrates seamlessly into any game engine.
Specifically, \cref{sec:vertex_based} and \cref{sec:triangle_based} introduce the new parametrization of \methodname to enable connectivity, \cref{sec:triangle_rasterization} details how triangles are projected into image space, \cref{sec:training_strategy} outlines the training strategy for achieving high mesh-based visual quality, and finally, \cref{sec:optimization} provides additional information on initialization and training losses. An overview of these steps is illustrated in \cref{fig:methodology}.

\subsection{Vertex-based representation}
\label{sec:vertex_based}

In previous Triangle Splatting methods \cite{Held2025Triangle-arxiv, Burgdorfer2025Radiant-arxiv}, each triangle is parameterized by three vertices $\mathbf{v}_i \in \mathbb{R}^3$, a color $\mathbf{c}$, a smoothness parameter $\sigma$, and an opacity $o$. Both $\sigma$ and $o$ are treated as free parameters, which prevents the triangles from being strictly opaque at the end of training. As a result, rendering the resulting triangle soup in a game engine leads to noticeable quality degradation.
Moreover, the formulation enforces triangles to remain isolated and unconnected. In regions of high density, many vertices lie close to each other and share similar color distributions, but cannot be merged or shared. Ideally, such cases should be represented by a single vertex connecting adjacent triangles, improving efficiency and structural consistency.

In \methodname, we therefore define a vertex set
\[
\mathcal{V} = \{\mathbf{v}_i \in \mathbb{R}^3 \mid i = 1,\dots,N\},
\]
with $N$ the total number of vertices. 
Each vertex is parameterized as $\mathbf{v}_i = (x_i, y_i, z_i, c_i, o_i),$
where $(x_i, y_i, z_i) \in \mathbb{R}^3$ denotes its 3D position, 
$\mathbf{c}_i \in \mathbb{R}^3$ the vertex color and $o_i \in [0,1]$ the vertex opacity, similar to~\cite{Son2024DMesh++-arxiv}.
After training, the opacity parameter can be discarded as each triangle will be opaque.

\subsection{Triangle representation}\label{sec:triangle_based}
A triangle is then defined by a triplet of vertex indices 
$\mathbf{T}_m = (i,j,k), \quad i,j,k \in \{1,\dots,N\}$,
which specifies the three vertices from $\mathcal{V}$ that build the triangle.  
The opacity of the triangle is computed as $o_{\mathbf{T}_m} = \min(o_i, o_j, o_k),$ and its color at a point inside the triangle is obtained by interpolating the vertex colors using barycentric coordinates 
$\mathbf{c}_{\mathbf{T}_m}(\lambda_i,\lambda_j,\lambda_k) 
= \lambda_i \mathbf{c}_i + \lambda_j \mathbf{c}_j + \lambda_k \mathbf{c}_k,$
with $\lambda_i + \lambda_j + \lambda_k = 1$ and $\lambda_i, \lambda_j, \lambda_k \geq 0$.
Triangles can now be easily combined by connecting them through shared vertices.
\noindent During training, the gradient of each triangle is propagated back to its corresponding vertices. Each vertex’s position, color, and opacity therefore receive the accumulated gradients from all connected triangles.

\subsection{Differentiable rasterization}\label{sec:triangle_rasterization}

The rasterization process begins by projecting each 3D vertex $\mathbf{v}_i$ onto the image plane using a standard pinhole camera model. The projection is defined by the intrinsic camera matrix $\mathbf{K}$ and the extrinsic parameters, i.e., rotation $\mathbf{R}$ and translation $\mathbf{t}$: $\mathbf{q}_i = \mathbf{K} \left( \mathbf{R}\mathbf{v}_i + \mathbf{t} \right)$,
where $\mathbf{q}_i \in \mathbb{R}^2$ denotes the 2D coordinates of the projected vertex in image space.
To determine the influence of a triangle on a pixel $p$, we use the same window function as used in Triangle Splatting \cite{Held2025Triangle-arxiv}.
The \emph{signed distance field} (SDF) $\phi$ of the 2D triangle in image space is given by:
$$
\phi(\mathbf{p}) = \max_{i\in \{1,2,3\}} L_i(\mathbf{p}),
\qquad
L_i(\mathbf{p}) = \mathbf{n}_i \cdot \mathbf{p} + d_i,
$$
where $\mathbf{n}_i$ are the unit normals of the triangle edges pointing outside the triangle, and $d_i$ are offsets such that the triangle is given by the zero-level set of the function $\phi$.
The signed distance field $\phi$ thus takes positive values outside the triangle, negative values inside, and equals zero on its boundary.
The window function $I$ is then defined as:
\begin{equation}\label{eq:indicator1}
\begin{aligned}
I(\mathbf{p}) &= \text{ReLU}\left( \frac{\phi(\mathbf{p})}{\phi(\mathbf{s})} \right)^{\sigma} \\
\text{such that} \quad
I(\mathbf{p}) &\begin{cases}
= 1 ~~\text{at the triangle incenter},\\
= 0 ~~\text{at the triangle boundary},\\
= 0 ~~\text{outside the triangle}.
\end{cases}
\end{aligned}
\end{equation}
with $\mathbf{s}\in\mathbb{R}^2$ be the \emph{incenter} of the projected triangle (\ie, the point inside the triangle with minimum signed distance).
$\sigma$ is a smoothness parameter that controls the transition between the interior and exterior of the triangle. As $\sigma \to 0$, the representation converges to a sharp, exact triangle, while larger values of $\sigma$ yield a smooth window function that gradually increases from zero at the boundary to one at the center.
Once the triangles are projected, the color of each image pixel~$\mathbf{p}$ is computed by accumulating contributions from all overlapping triangles, in depth order given:

\begin{equation}
\label{equ:contr}
    C(\mathbf{p}) = \sum_{n=1}^N \mathbf{c}_{T_n} o_{T_n} I(\mathbf{p}) \left( \prod_{i=1}^{n-1} \left(1 - o_{T_i} I(\mathbf{p})\right) \right),
\end{equation}
At the end of training, \cref{equ:contr} simplifies to
$C(\mathbf{p}) = \mathbf{c}_{T_n} I(\mathbf{p})$,
so that only a single evaluation per pixel is required, significantly accelerating the rendering process.

\subsection{Training strategy}\label{sec:training_strategy}

The training strategy is the most important part for obtaining a high-visual quality with opaque primitives.
The first challenge is that during optimization, it is essential that gradients can be back-propagated. 
With only solid and opaque triangles (low $\sigma$ and high $o$), gradients would vanish, making optimization impossible. 
To ensure gradient flow, we allow triangles during training to be smooth (soft transition from inside the triangle to outside) and semi-transparent. 
The transition from semi-transparent and soft triangles to solid and opaque ones is crucial for achieving high visual quality in the final reconstruction.
Secondly, pruning triangles becomes increasingly important. Since opacity is manually increased during training, triangles can no longer remain hidden by staying transparent. By the end of training, all triangles become opaque, even if they degrade visual quality. It is therefore essential to filter out such triangles during optimization to avoid artifacts in the final result.
In the following, we first describe how $\sigma$ and opacity are optimized during training to ensure gradient flow while still yielding fully opaque triangles at convergence. We then introduce our pruning and densification strategy designed to achieve high visual quality.

\mysection{Triangles smoothness.}
Unlike Triangle Splatting, where $\sigma$ is freely optimized and defined per triangle, we initialize it to $1.0$ (corresponding to a linear transition from the incenter to the boundary) and treat it as a single parameter shared across all triangles. 
During training, we gradually anneal $\sigma$ to $0.0001$, providing strong gradient flow in the early stages and converging to sharp, well-defined triangles at the end.  

\mysection{Triangles opacities.}
In traditional splatting methods, opacity is mapped to the $[0,1]$ domain using a Sigmoid activation.
In our approach, however, we enforce a gradually increasing mapping domain by redefining the activation as
$\text{opacity}(x) = O_t + (1 - O_t)\,\sigma(x)$,
where $O_t$ denotes the opacity floor. 
After the first $5\text{k}$ training iterations, we update the mapping to constrain opacities to $[O_t, 1]$, and anneal $O_t$ over the course of training until all triangles become opaque.
Unlike the smoothness parameter $\sigma$, opacity values remain free variables throughout optimization but is constrained to lie within $[O_t, 1]$, instead of fixing them to a constant. 
After training,  $\sigma$ and the opacity can be discarded.

\begin{figure}[t]
\centering

\zoomin{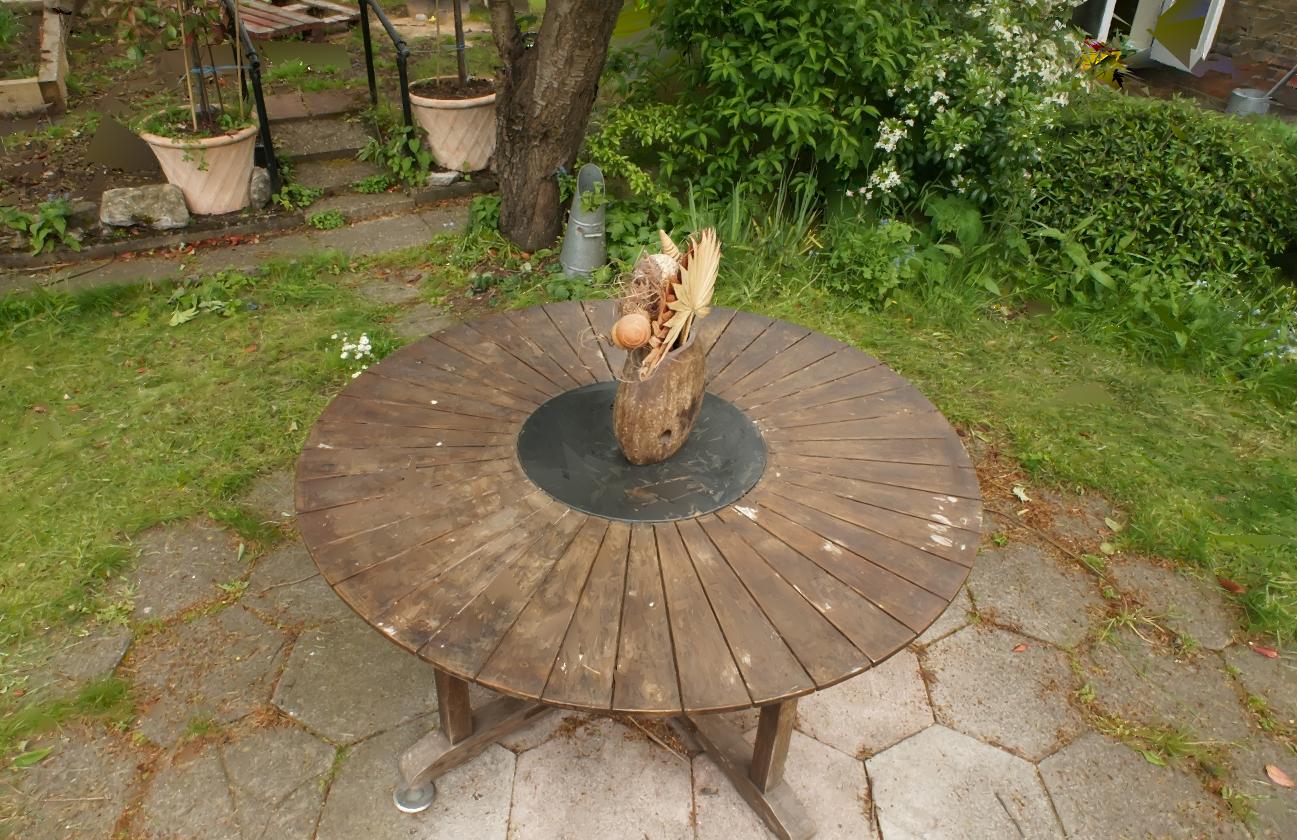}{0.1\mytmplen}{0.53\mytmplen}{0.81\mytmplen}{0.19\mytmplen}{1.5cm}{\mytmplen}{5.8}{red} \hfill
\zoomin{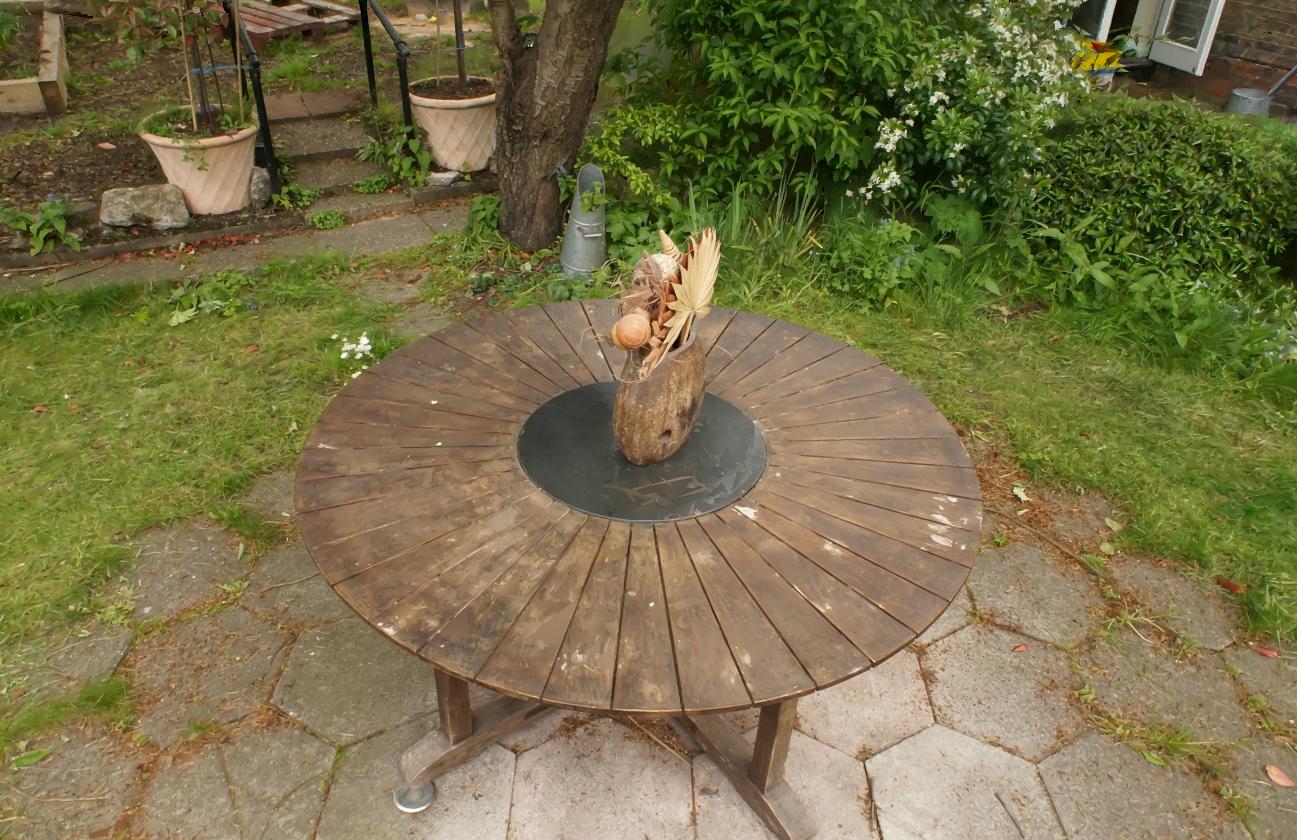}{0.1\mytmplen}{0.53\mytmplen}{0.81\mytmplen}{0.19\mytmplen}{1.5cm}{\mytmplen}{5.8}{red}
\captionof{figure}{ \small
\textbf{Hard pruning step.}
We eliminate unnecessary triangles (right) that would otherwise remain in the scene (left).
\label{fig:pruning}
}
\end{figure}

\mysection{Pruning.}
In traditional splatting methods, pruning primarily serves to reduce memory consumption and improve rendering speed. In the worst case, a primitive simply converges to a low opacity and becomes invisible in the final 3D representation.
In our setting, however, pruning plays a far more critical role. During training, our representation gradually converges to opaque triangles. Unlike in traditional splatting, our triangles cannot hide behind low opacities, since this would contradict our training objective. Consequently, it is essential to employ an effective pruning strategy to ensure that redundant triangles are removed before they become solid and opaque, as otherwise they would persist in the final scene representation and significantly degrade visual quality.
During the first $5k$ iterations, opacity is not restricted, allowing important triangles to converge toward high opacities while less important ones naturally decay to low opacity.
After $5k$ iterations, we apply a \textit{hard pruning step} that removes all triangles with opacity below a threshold $T_o \approx 0.2$. This eliminates roughly $70\%$ of the triangles and vertices, but ultimately improves the final visual quality. Using a lower threshold $T_o$ reduces the immediate drop in quality, but in later stages, when triangles become opaque, many redundant primitives remain. \cref{fig:pruning} illustrates the effect of this hard pruning step.
After $5k$ iterations, pruning continues. However, removing triangles solely based on low opacity $o$ becomes ineffective, since opacities are explicitly pushed toward high values and our mapping function no longer permits low opacities.
To identify triangles that have little influence or are completely occluded, we instead compute the maximum volume rendering weight $T \cdot o$ (with $T$ denoting transmittance and $o$ opacity) for each triangle during rasterization. Triangles whose maximum weight falls below a threshold $\tau_\text{prune}$ across all training views are discarded. Toward the end of training, as triangles become increasingly opaque and sharp, many triangles are pruned because they lie behind others and no longer contribute to the rendering.
Vertices are pruned once they are no longer connected to any triangle.
A detailed analysis of the pruning impact is provided in \cref{sec:pruning_strategy}.

\mysection{Densification.}
Since the initial representation does not contain enough triangles and vertices to achieve high visual quality, we adopt a probabilistic MCMC-based framework~\cite{Kheradmand20243DGaussian} to progressively introduce additional triangles. At each densification step, candidate triangles are selected by sampling from a probability distribution constructed directly from their opacity $o$ using Bernoulli sampling. 
New triangles and vertices are then generated through \emph{midpoint subdivision}: the midpoints of the three edges of a selected triangle are connected, splitting it into four smaller triangles. The new midpoints are added to the vertex set $\mathcal{V}$ and assigned the average color and opacity of their two adjacent vertices.
This operation preserves both the total area and the spatial extent of the original primitive, and ensures that the subdivided triangles remain connected.

\subsection{Optimization}
\label{sec:optimization}

\begin{table*}[t!]
	\small
\resizebox{\linewidth}{!}{
  \tabcolsep=0.07cm
\begin{tabular}{l|ccc|cccc|cccc}
	Dataset & & & \multicolumn{4}{c}{Mip-NeRF360 Dataset} & \multicolumn{4}{c}{Tanks \& Temples} \\
	Method|Metric & E2E mesh & E2E colored & Mesh Con.
	& $PSNR^\uparrow$ & $LPIPS^\downarrow$     & $SSIM^\uparrow$  & \#Verts$^\downarrow$ & $PSNR^\uparrow$ & $LPIPS^\downarrow$     & $SSIM^\uparrow$  & \#Verts$^\downarrow$ \\
	\midrule 
    3DGS\cite{Kerbl20233DGaussian} & / & / & / & 27.21 & 0.214 & 0.815 & / & 23.14 & 0.183 & 0.841 & / \\
\midrule
  2DGS~\cite{Huang20242DGaussian} & \xmark & \xmark & \cmark & 15.36 & 0.474  & 0.498 & \best 2M  & 14.23 &  0.485  & 0.569 & 16M  \\
  GOF~\cite{Yu2024Gaussian} & \xmark & \xmark & \cmark & 20.78 & 0.465 & 0.573 & 33M & \best 21.69 & 0.326 & 0.690 & 12M \\
  RaDe-GS~\cite{Zhang2024RaDeGS-arxiv} & \xmark & \xmark & \cmark & 23.56 & 0.361 & 0.668 & 31M & 20.51 & 0.344 & 0.659 &10M \\
  MiLo~\cite{Guedon2025MILo-arxiv} & \cmark & \xmark & \cmark & 24.09 & 0.323 & 0.688 & 7M &  21.46 & 0.348 & 0.706 & 4M \\
  Triangle Splatting~\cite{Held2025Triangle-arxiv} $\dagger$ & \cmark & \cmark & \xmark & 21.05 & 0.462 & 0.558 &	3M  & 17.27 & 0.402 & 0.600 & 6M \\
\midrule
    Ours & \cmark & \cmark & $\sim$ & \best 25.21  & \best 0.294 & \best 0.742  & \best 2M & 20.91 & \best 0.249 & \best 0.773  & \best 2M  \\
\end{tabular}
	}
\caption{\textbf{Mesh-based novel view synthesis on the Mip-NeRF360 dataset.} \emph{E2E mesh} indicates whether a method directly produces a mesh. 
\emph{E2E colored} denotes whether the mesh is already colored or requires post-processing. 
\emph{Mesh Con.} specifies whether the reconstructed mesh consists of a connected component.  $\sim$ stands for semi-connectivity.
$\dagger$ with only opaque triangles.
}
 \label{tab:comparisons_main_table}
\end{table*}

\mysection{Initialization.} 
Our method starts from a set of images and their corresponding camera parameters, calibrated via SfM~\cite{Schonberger2016Structure}, which also provides a sparse point cloud.  
We apply 3D Delaunay triangulation on this point cloud to obtain a tetrahedralization, from which we extract all unique triangles. This initial mesh provides a well-connected set of triangles that serves as the starting point.

\mysection{Parameters \& losses.}
We optimize the 3D vertex positions $\mathbf{v}_i$, opacity $o_i$, and spherical harmonic color coefficients $\mathbf{c_i}$ of all vertices by minimizing the rendering error from the given posed views.
Our training loss combines the photometric $\mathcal{L}_1$ and $\mathcal{L}_{\text{D-SSIM}}$ terms~\cite{Kerbl20233DGaussian}, the opacity loss $\mathcal{L}_o$~\cite{Kheradmand20243DGaussian}, and normal $\mathcal{L}_n$ losses. 
The final loss $\mathcal{L}$ is given by:
\begin{equation}
    \mathcal{L} = (1 - \lambda) \mathcal{L}_1 + \lambda \mathcal{L}_{\text{D-SSIM}} + \beta_1 \mathcal{L}_o + \beta_2 \mathcal{L}_n \, .
\end{equation}
For the normal loss $\mathcal{L}_n$, we supervise using a normal estimation model \cite{Hu_2024}.

\mysection{Anti-aliasing.} To mitigate aliasing artifacts, we render at $s\times$ the target resolution and then downsample to the final resolution using area interpolation, which averages over input pixel regions and acts as an anti-aliasing filter.

\section{Experiments}
\label{sec:experiments}

\mysection{Implementation details.}
For all experiments, we set the spherical harmonics to degree 3, which yields 51 parameters per vertex (48 from the SH coefficients and 3 from the vertex position) and 3 parameters per triangle. In comparison, a single Gaussian in 3DGS requires 59 parameters.

\mysection{Task.}
Prior work \cite{Kerbl20233DGaussian, Mildenhall2020NeRF-eccv} evaluates reconstruction quality by comparing synthesized and ground-truth images, independent of whether the representation is implicit, explicit, or semi-transparent. We instead focus on the visual fidelity of the reconstructed mesh itself, using only opaque triangles.
To this end, we follow MiLo \cite{Guedon2025MILo-arxiv} and adopt the task of \textit{Mesh-Based Novel View Synthesis}, which assesses how well reconstructed meshes represent complete scenes.

\mysection{Baselines.} 
We compare against Triangle Splatting (restricted to solid and opaque triangles), and meshes derived from MiLo~\cite{Guedon2025MILo-arxiv}, 2DGS~\cite{Huang20242DGaussian}, Gaussian Opacity Fields (GOF) \cite{Yu2024Gaussian}, and RaDe-GS \cite{Zhang2024RaDeGS-arxiv}, following the protocol of~\cite{Guedon2025MILo-arxiv}.
For MiLo, surface mesh extraction is integrated into the optimization itself, so no additional post-processing is required to obtain a mesh output. In contrast, 2DGS, GOF, and RaDe-GS rely on mesh extraction after optimization.
Nevertheless, all these methods require an additional post-processing stage to color the mesh, achieved by training a neural color field for 5k iterations (see MiLo for details~\cite{Guedon2025MILo-arxiv}).
For Triangle Splatting, we adopt the \textit{game engine training strategy}, which produces opaque triangles, and no additional post-processing is needed.
For reference, we also compare against the original 3D Gaussian Splatting \cite{Kerbl20233DGaussian} to highlight the contrast between radiance-field rendering quality and mesh-based novel view synthesis.
For our method, the final output consists of opaque, colored triangles, making it immediately compatible with mesh renderers without requiring any post-processing.

\mysection{Datasets \& Metrics.}
We compare our method to concurrent approaches on the standard benchmarks Mip-NeRF360~\cite{Barron2022MipNeRF360} and Tanks and Temples (T\&T)~\cite{Knapitsch2017Tanks}.
We evaluate the visual quality using standard metrics: SSIM, PSNR, and LPIPS.
We report training time on an NVIDIA A100 and the total number of used vertices.

\begin{figure*}[t]
\centering
\setlength{\mytmplen}{0.18\linewidth}
\resizebox{\linewidth}{!}{ 
\begin{tabular}{c@{\hskip 0.05in}c@{\hskip 0.05in}c@{\hskip 0.05in}c}
    
    & \makebox[\mytmplen]{Ground Truth} &
      \makebox[\mytmplen]{\textbf{\methodname}} &
      \makebox[\mytmplen]{Triangle Splatting$\dagger$} 
   \\

    \rotatebox{90}{\parbox{2.2cm}{\centering Bicycle}} &
    \zoomin{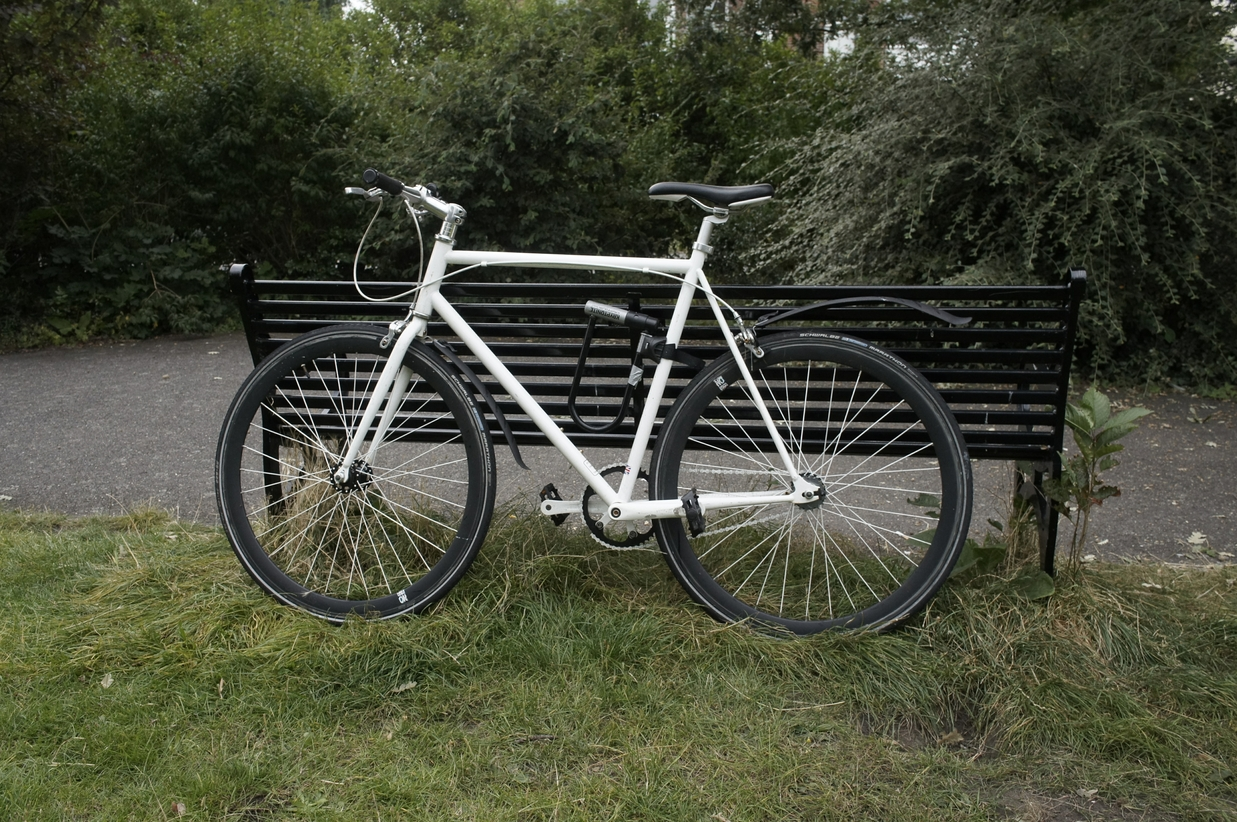}{0.27\mytmplen}{0.23\mytmplen}{0.846\mytmplen}{0.157\mytmplen}{0.95cm}{\mytmplen}{3.5}{red}
    & \zoomin{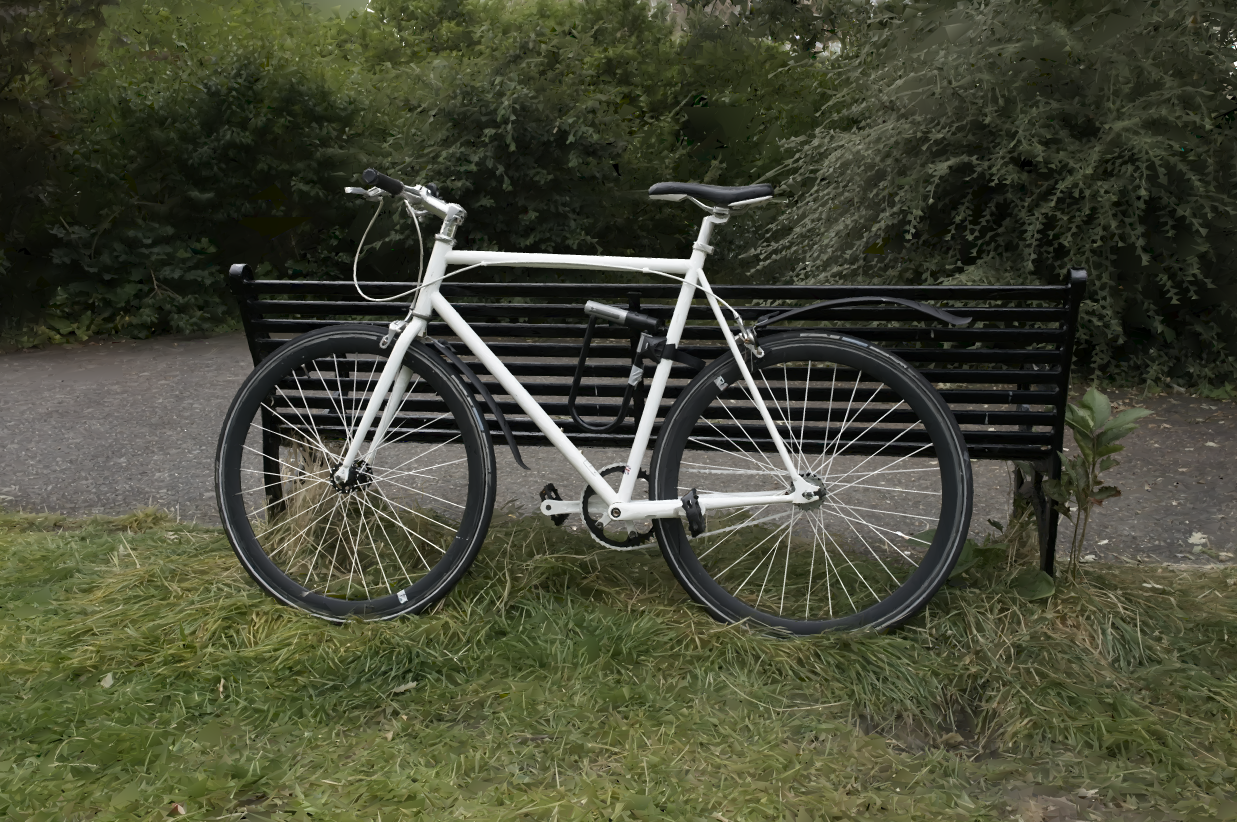}{0.27\mytmplen}{0.23\mytmplen}{0.846\mytmplen}{0.157\mytmplen}{0.95cm}{\mytmplen}{3.5}{red}  &
    \zoomin{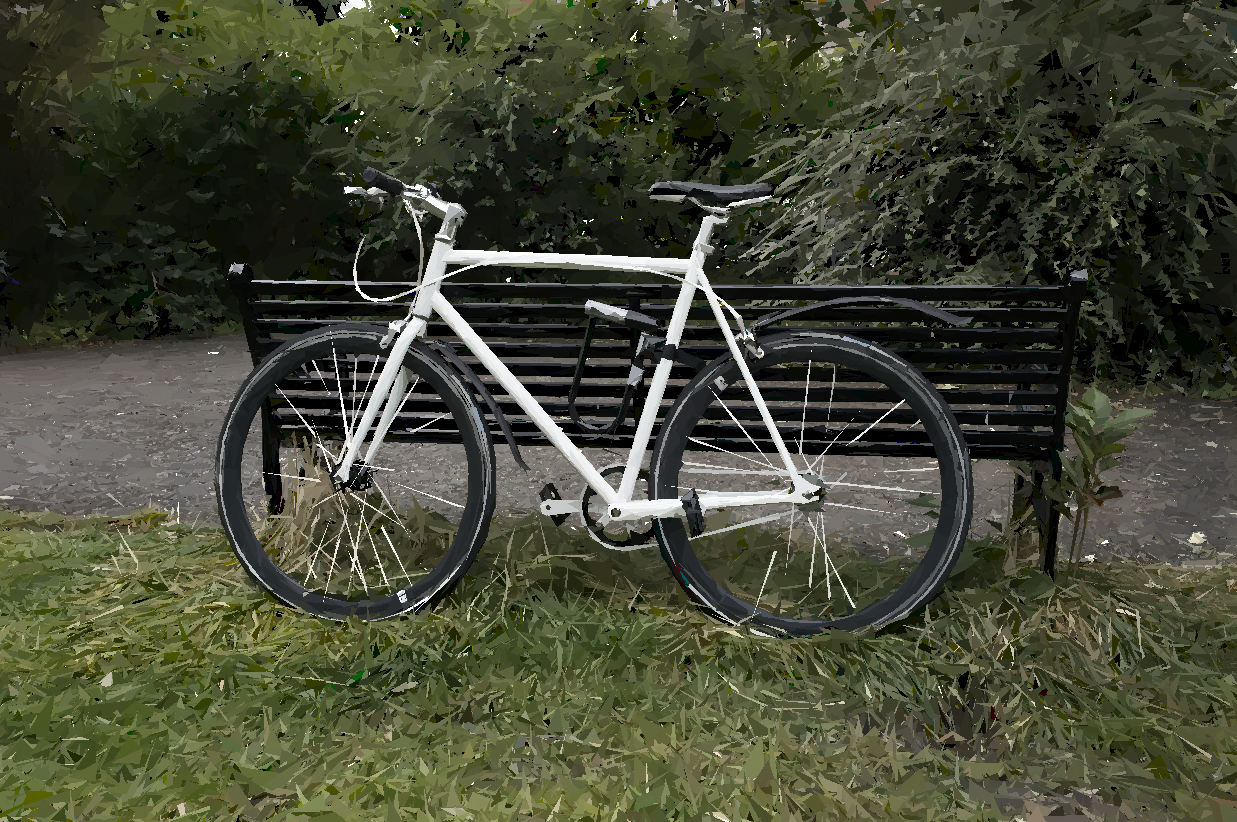}{0.27\mytmplen}{0.23\mytmplen}{0.846\mytmplen}{0.157\mytmplen}{0.95cm}{\mytmplen}{3.5}{red}  
    \\

    \rotatebox{90}{\parbox{2.2cm}{\centering Garden}} &
    \zoomin{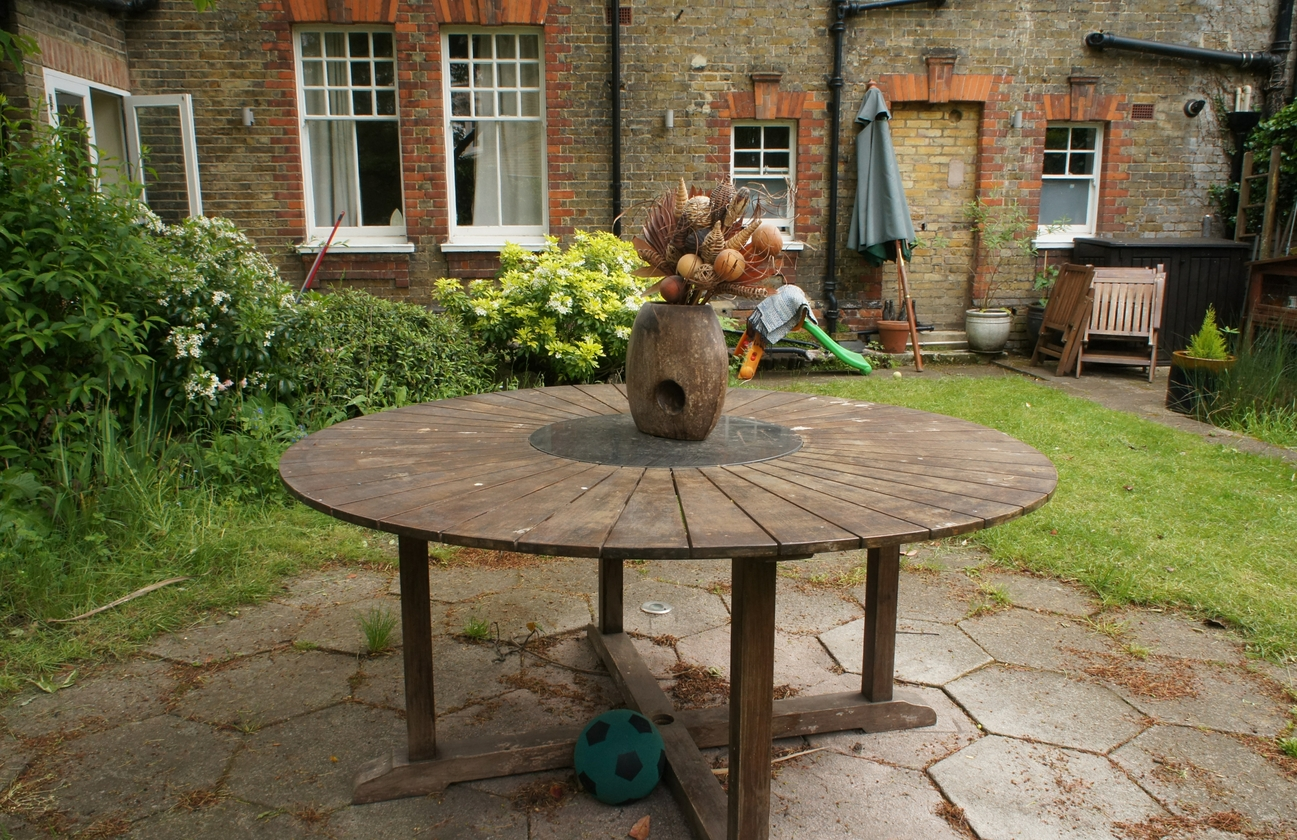}{0.54\mytmplen}{0.48\mytmplen}{0.846\mytmplen}{0.157\mytmplen}{0.95cm}{\mytmplen}{4.5}{red}  

   &  \zoomin{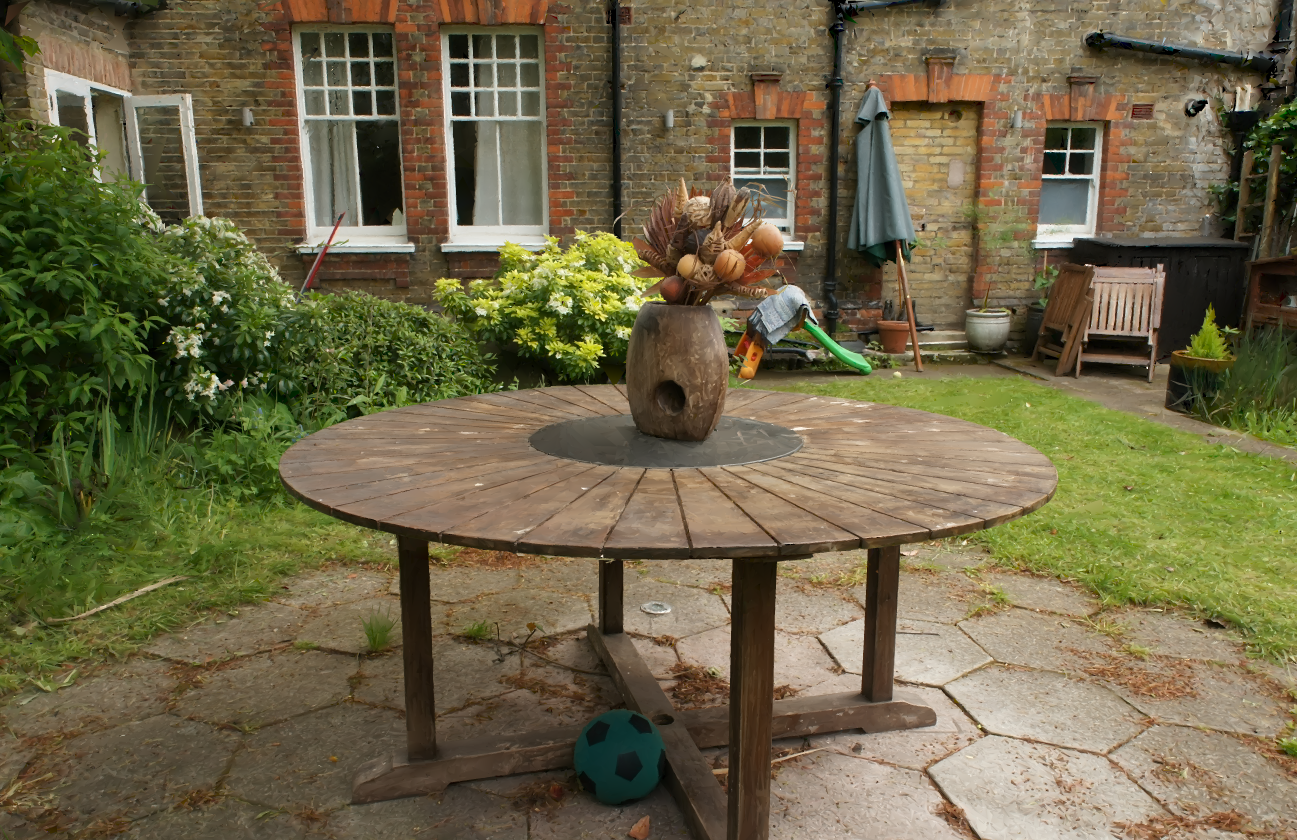}{0.54\mytmplen}{0.48\mytmplen}{0.846\mytmplen}{0.157\mytmplen}{0.95cm}{\mytmplen}{4.5}{red}  & \zoomin{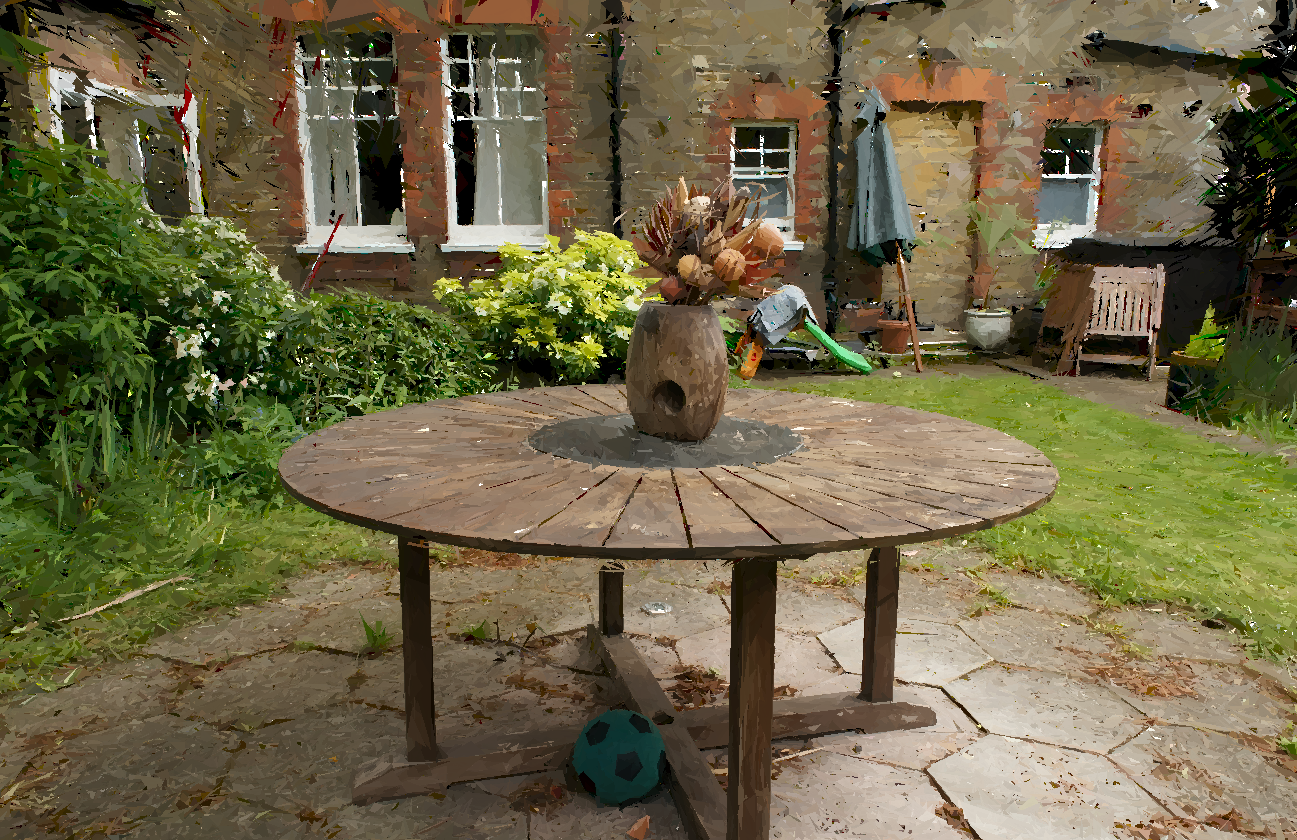}{0.54\mytmplen}{0.48\mytmplen}{0.846\mytmplen}{0.157\mytmplen}{0.95cm}{\mytmplen}{4.5}{red}  
   \\

    \rotatebox{90}{\parbox{2.2cm}{\centering Counter}} &
    \zoomin{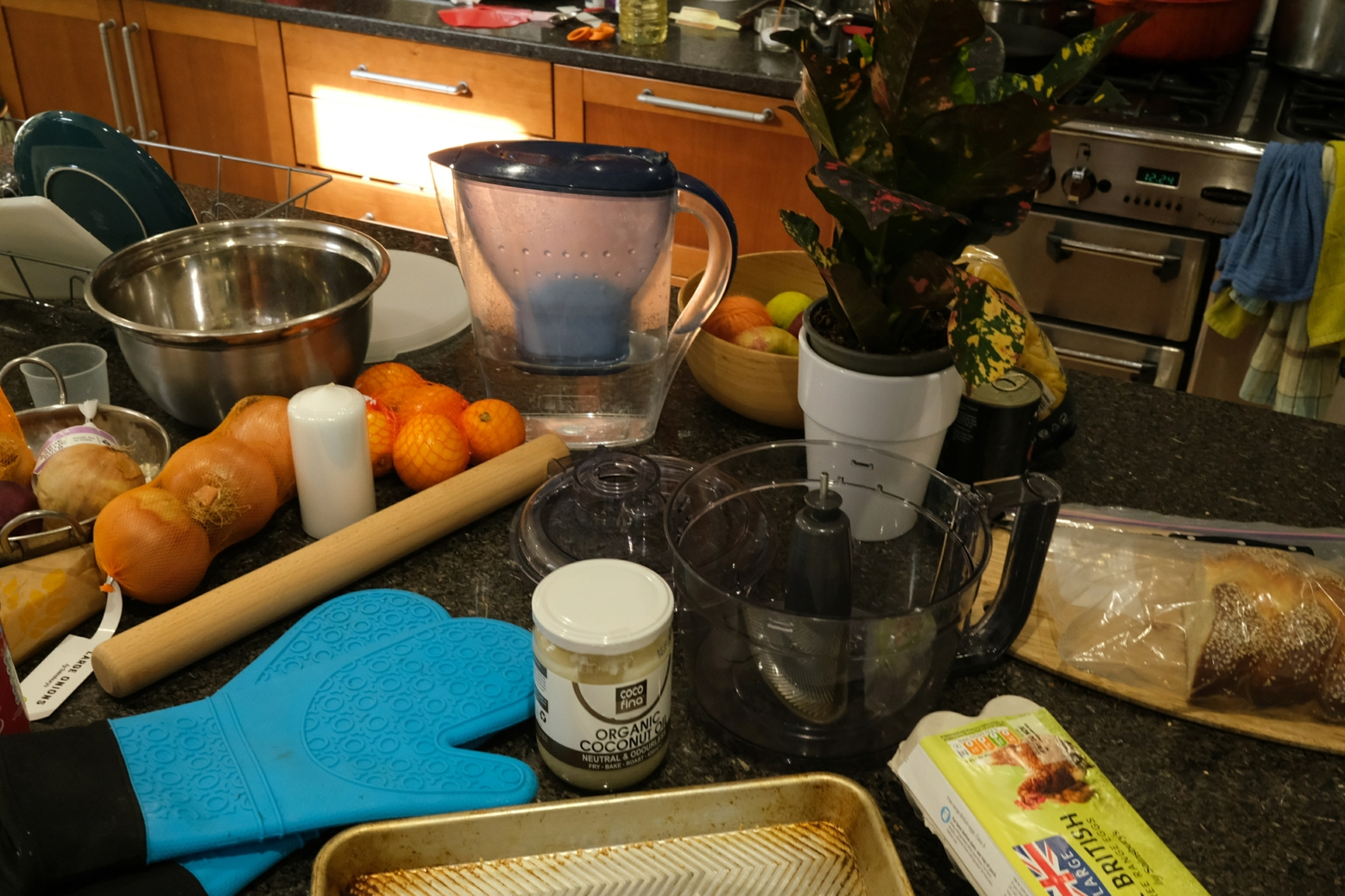}{0.46\mytmplen}{0.13\mytmplen}{0.846\mytmplen}{0.157\mytmplen}{0.95cm}{\mytmplen}{4.5}{red}    
   & \zoomin{images/qualitative_results/ts+_counter_view_7.png}{0.46\mytmplen}{0.13\mytmplen}{0.846\mytmplen}{0.157\mytmplen}{0.95cm}{\mytmplen}{4.5}{red}  &  \zoomin{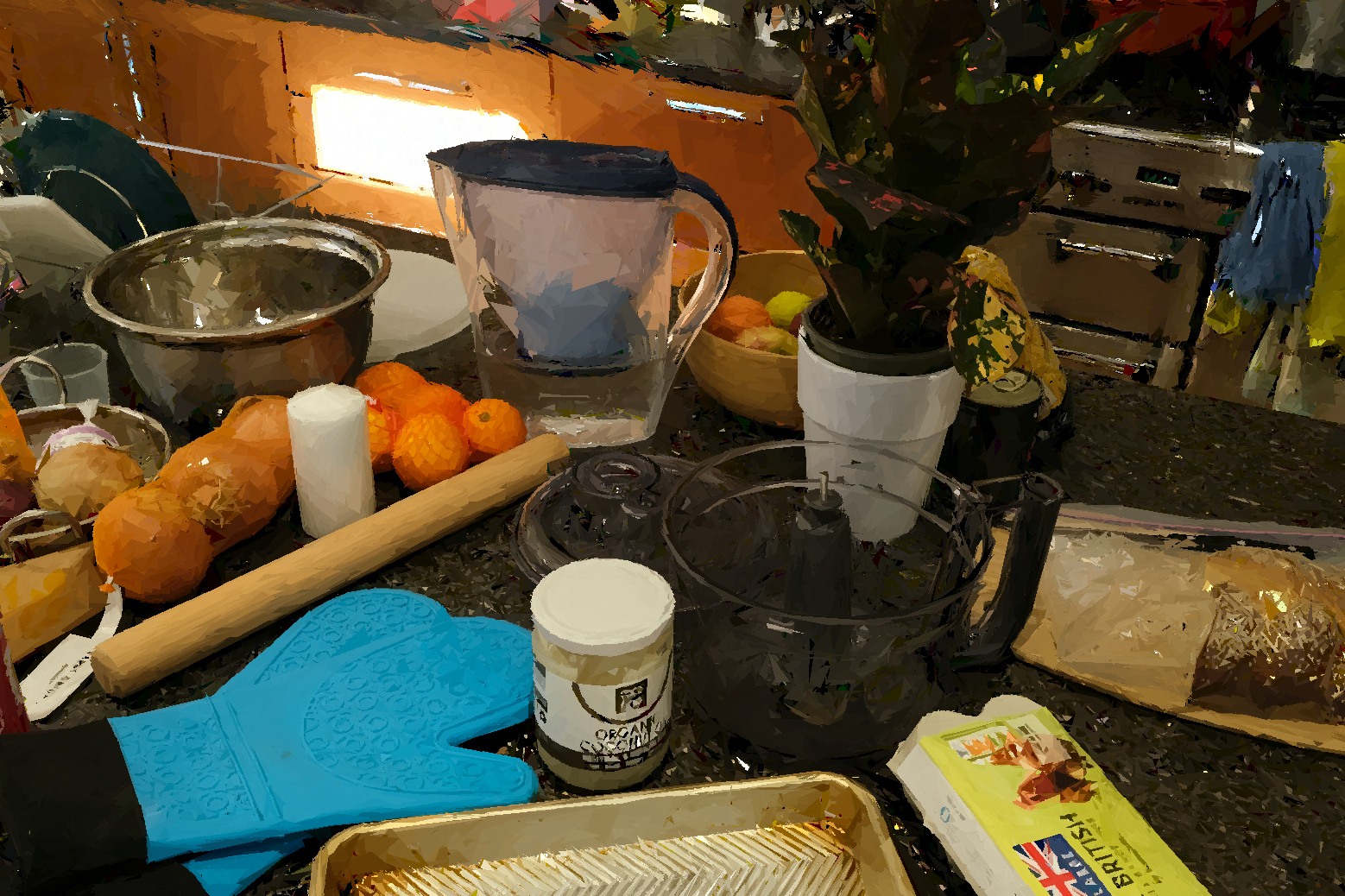}{0.46\mytmplen}{0.13\mytmplen}{0.846\mytmplen}{0.157\mytmplen}{0.95cm}{\mytmplen}{4.5}{red}  
    \\

\end{tabular}
}

\caption{\small \textbf{Qualitative results.} 
Comparison of our method with Ground Truth and opaque Triangle Splatting~\cite{Held2025Triangle-arxiv}. Our approach produces renderings that are closer to the ground truth, with sharper details and more faithful recovery of fine structures. $\dagger$ Opaque version of Triangle Splatting.
}
\label{fig:qualityresults}
\end{figure*}

\subsection{Mesh-Based Novel View Synthesis}

\cref{tab:comparisons_main_table} reports quantitative results on the Mip-NeRF360 and Tanks \& Temples datasets for mesh-based novel view synthesis.
\methodname consistently outperforms all concurrent methods across all metrics.
Compared to 2DGS and Triangle Splatting, our method uses a similar number of vertices yet achieves a 4–10 dB higher PSNR.
GOF, RaDe-GS, and MiLo require 2–10× more vertices while still obtaining lower PSNR and SSIM and higher LPIPS.
In terms of LPIPS~(the metric that best correlates with human visual perception), \methodname significantly outperforms all concurrent methods.
Furthermore, 2DGS, GOF, and RaDe-GS require two additional post-processing steps after training: first extracting a mesh, and then coloring it. MiLo directly outputs a surface mesh after training, but still relies on a post-processing stage to texture the mesh. These extra steps limit the practicality of such methods, and increase the overall pipeline complexity.
In contrast, our method directly produces colored, opaque triangles that are immediately compatible with any game engine, without requiring additional steps. 
Although 2DGS, GOF, RaDe-GS, and MiLo produce watertight and fully connected meshes, our representation is still sufficient for key downstream uses such as collision-based physical simulation or walkable environments in interactive games.
Furthermore, Simplicits~\cite{Modi2024Simplicits} shows that elastic simulation can be performed directly on non-watertight, even mesh-free representations. \methodname representation enables the same downstream applications without requiring a fully watertight mesh.
\cref{fig:qualityresults} shows qualitative results, demonstrating that \methodname produces sharp renderings and reconstructs fine details.

\mysection{Training speed.} The training speed of \methodname is considerably higher than that of concurrent approaches. On the Mip-NeRF360 dataset, \methodname trains in 39 minutes, and on the T\&T dataset in 25 minutes. By comparison, MiLo requires around 45 minutes on T\&T and up to 2 hours on Mip-NeRF360. While Triangle Splatting achieves shorter training times (17 minutes on Mip-NeRF360 and 20 minutes on T\&T), this comes at the cost of substantially lower visual quality.

\subsection{Downstream applications.}

\begin{figure}[ht]
\centering
\includegraphics[width=0.48\linewidth]{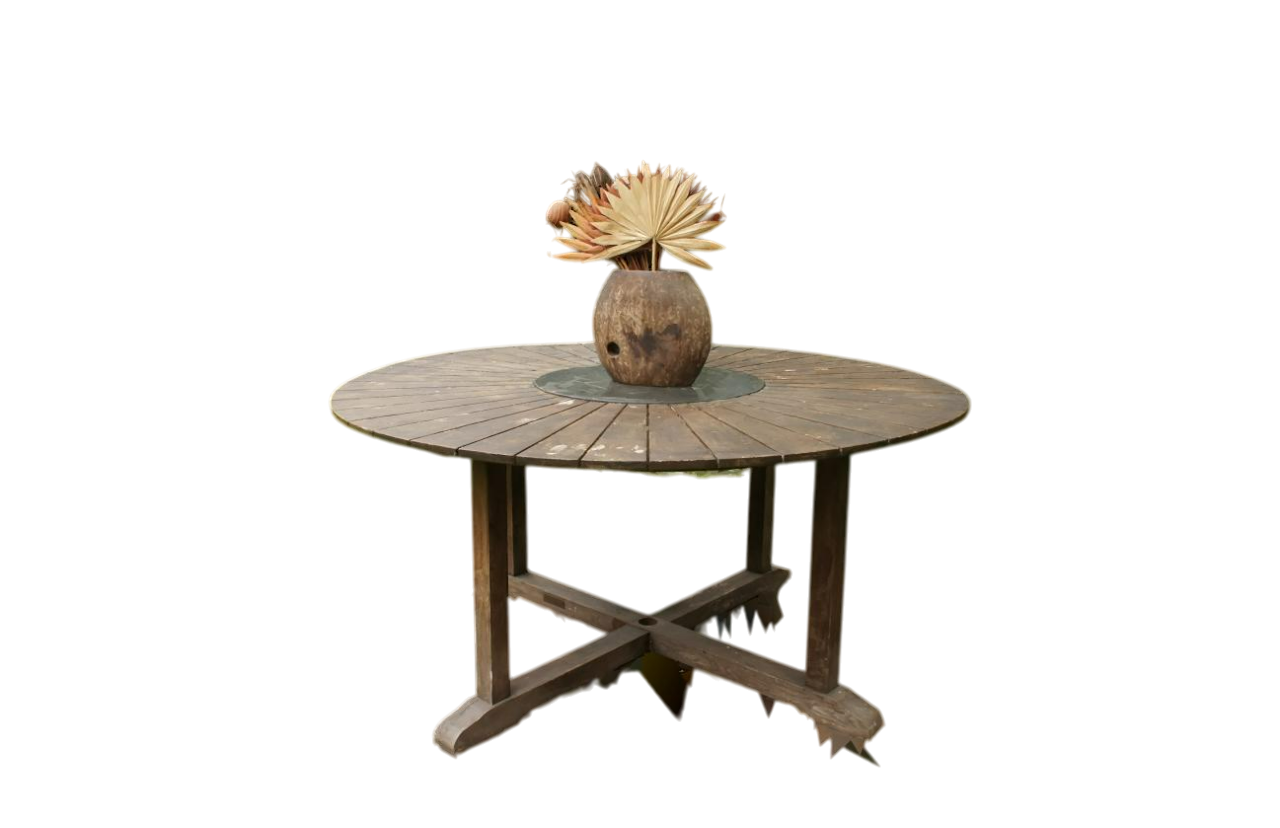}
\includegraphics[width=0.48\linewidth]{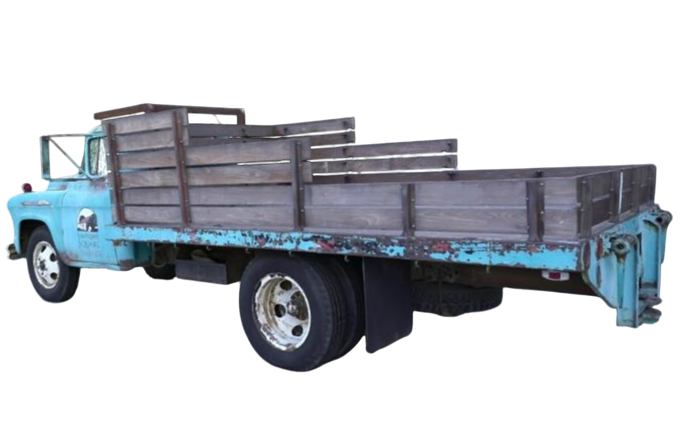}
\caption{\small
\textbf{Straightforward extraction of objects.} With \methodname, objects can be extracted or removed from a scene.
}
\label{fig:extraction}
\end{figure}

\mysection{Removing or extracting objects.}
Current 3D Gaussian Splatting methods for object extraction or removal \cite{Cen2025Segment, Ye2024Gaussian} face a fundamental challenge: a single pixel is influenced by the accumulated contributions of many primitives, rather than being assigned to a single one. This makes it non-trivial to decide whether a primitive belongs to a given object. To address this, prior work learns object associations during optimization \cite{Cen2025Segment, Ye2024Gaussian}.
In contrast, \methodname requires no additional optimization during training. Since each pixel is determined by exactly one triangle, mapping objects from image space to 3D space becomes straightforward: given a 2D mask of an object, all triangles contributing to pixels within that mask are directly identified as part of the object. By iterating over all training views, we recover the complete set of triangles belonging to the object. 
The object masks are generated using Segment Anything 2 \cite{Ravi2025SAM2}, which enables the selection of single or multiple objects that can then be removed or extracted from a scene with minimal additional processing.
\cref{fig:extraction} presents qualitative examples of two objects extracted from the Mip-NeRF360 and Tanks \& Temples datasets.

\mysection{Physical simulation and walkable environments.}
As our representation contains no transparent or nearly invisible primitives, the triangles may directly be interpreted as hard surfaces for the purpose of physics simulation.
We demonstrate this by using an off-the-shelf non-convex mesh collider implementation, specifically the one provided in the Unity game engine.
With no post-processing, our mesh can be loaded into a Unity scene and used for physics interaction with dynamic objects and characters.
Additional visualizations are available on our project page. We also provide the PLY files for \textit{Garden}, \textit{Bicycle}, and \textit{Truck}, along with a \textit{Unity} project for exploring physics-based interactions and walkable scenes.

\subsection{Ablation study}

\mysection{Pruning and training strategy.}\label{sec:pruning_strategy}
An effective pruning strategy is crucial for achieving high visual quality with opaque primitives. \cref{table:ablation_losses} summarizes the impact of the two pruning strategies on \methodname.
Without the hard pruning step at iteration $5k$, visual quality degrades noticeably. Unnecessary triangles remain in the scene and can no longer be removed.
Similarly, when relying only on opacity as the pruning criterion, triangles can no longer be removed once the opacity floor has increased, since none of them are allowed to fall below the threshold. In contrast, using the blending weight as the criterion still enables pruning: triangles hidden behind others receive progressively lower transmittance as the front triangles become solid and opaque, and their blending weight eventually drops below the pruning threshold.
Instead of initializing with soft triangles (i.e., $\sigma = 1.0$) and gradually converging towards solid triangles, we ablate the case where training starts directly with solid triangles, such that gradients can only be propagated through opacity. In this setting, performance and visual quality drop significantly, as optimization barely progresses due to vanishing gradients.

\begin{table}[h]
\centering
\resizebox{0.9\linewidth}{!}{%
\begin{tabular}{c|ccc}
 Method  & PSNR~$\uparrow$ & LPIPS~$\downarrow$ & SSIM~$\uparrow$ \\
\midrule
Baseline & 25.28 & 0.289 & 0.751 \\
w/o hard pruning step  & -0.46 & +0.029  & -0.025  \\
w/o blending weight pruning & -0.51 & +0.069 & -0.045 \\
w/o sigma decay & -6.84 & +0.243 & -0.282 \\
\end{tabular}%
}
\caption{\small
\textbf{Pruning and training strategy ablated on Mip-NeRF360.} 
We isolate the impact of each design choice by removing them individually.
}
\label{table:ablation_losses}
\end{table}

\begin{table}[ht]
\centering
\resizebox{0.9\linewidth}{!}{%
\begin{tabular}{cc|ccc}
 $\sigma$ & $o$ & PSNR~$\uparrow$ & LPIPS~$\downarrow$ & SSIM~$\uparrow$ \\
\midrule
Hard & Opaque & 25.28 & 0.289 & 0.751 \\
Soft & Opaque &  +0.31 & -0.018 & +0.013 \\
Hard & Free & +1.26 & -0.048 & +0.032 \\
Soft & Free & \textbf{+1.34}  & \textbf{-0.054} & \textbf{+0.038} \\
\end{tabular}%
}
\caption{\small
\textbf{Relative improvements for different $\sigma$ and opacity settings.} 
Here, \emph{soft} denotes $\alpha=0.1$, \emph{hard} corresponds to a sharp transition, \emph{opaque} enforces a fixed opacity of $1$, and \emph{free} allows opacity to be optimized.  
}
\label{table:ablation_sigma_opacity}
\end{table}

\mysection{Soft and semi-transparent triangles.}
\cref{table:ablation_sigma_opacity} shows how visual quality is affected when using solid and semi-transparent triangles, soft and opaque triangles, or soft and semi-transparent triangles, compared to the baseline with solid and opaque triangles.
While higher final $\sigma$ values yield slight improvements in visual quality, the gains are significantly smaller than those obtained by allowing opacity to remain optimizable rather than forcing it to be fully opaque. 
When the only objective is to maximize visual quality, training with free opacity yields the best results, as semi-transparent triangles can be effectively blended and their colors accumulated to produce higher-quality renderings. However, in game engines, rendering \textit{efficiency} depends on skipping sorting. For this reason, enforcing fully opaque triangles is essential for achieving the fastest rendering performance.  

\mysection{Triangle connectivity.}
During optimization, we split triangles and connect subsets of them, but the optimization process does not strictly enforce full connectivity. Mainly because of pruning, connectivity is only partially preserved. On average, each vertex is connected to 1.5 triangles. Overall, 80\% of the triangles are connected to at least one other triangle, with some triangles connected to as many as six.

\begin{figure}[t]
\setlength\mytmplen{0.48\linewidth}
\centering
\zoomin{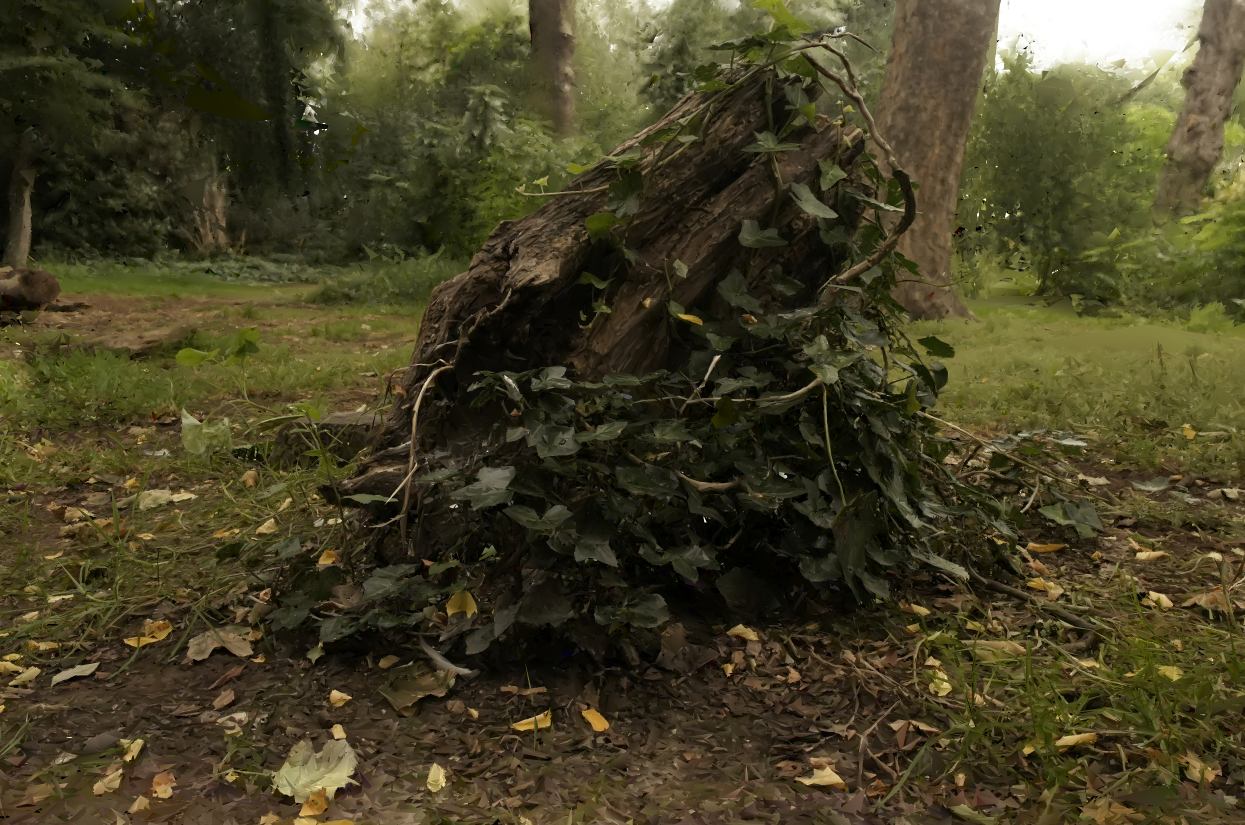}{0.3\mytmplen}{0.6\mytmplen}{0.81\mytmplen}{0.19\mytmplen}{1.5cm}{\mytmplen}{3.5}{red} \hfill
\zoomin{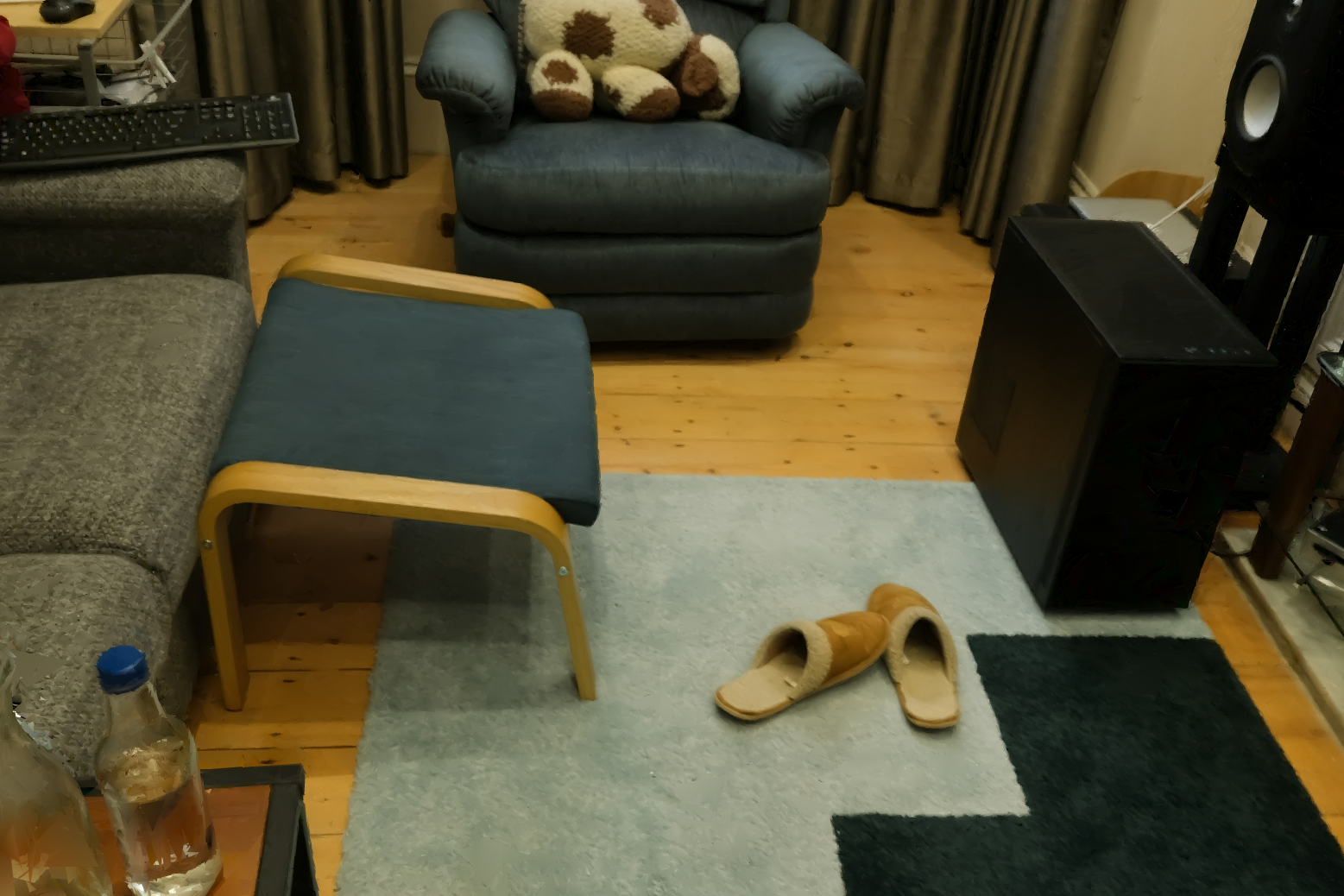}{0.1\mytmplen}{0.1\mytmplen}{0.81\mytmplen}{0.19\mytmplen}{1.5cm}{\mytmplen}{3.5}{red} 
\captionof{figure}{\small
\textbf{Limitations.} Accurately recovering backgrounds (left), particularly under limited viewpoints, and handling transparent objects (right) remain challenging.
\label{fig:limitations}
}
\end{figure}

\mysection{Limitations.}
\methodname achieves high visual quality and accurate reconstruction in regions where the initial point cloud is dense. In contrast, background areas with sparse coverage still exhibit incomplete geometry and reduced fidelity. 
Moreover, when moving outside the orbit of training views, the visual quality degrades. While the softness and opacity of Gaussian-based approaches may still provide slightly plausible results in such cases, our use of opaque triangles makes the artifacts more pronounced.
Future work could address those limitation by initializing with a more complete point cloud, or by incorporating alternative additional representation such as a triangulated sky dome.
Finally, transparent objects such as glasses or bottles remain difficult to represent using only opaque triangles, as illustrated in \cref{fig:limitations}.
\section{Conclusion}
\label{sec:conclusion}

We introduced \methodname, a differentiable rendering framework that optimizes opaque triangles with shared-vertex connectivity. With a tailored training strategy for opaque primitives, our method achieves state-of-the-art performance in mesh-based novel view synthesis while remaining efficient to train. Unlike Gaussian-based approaches, the resulting representation is immediately compatible with game engines, enabling downstream applications such as relighting, physical simulation, and interactive walkthroughs. \methodname bridges radiance field optimization with traditional graphics pipelines, paving the way for practical integration of radiance field representations into interactive VR applications, game engines, and simulation frameworks

\clearpage
{
    \small
    \bibliographystyle{plainnat} 
    \bibliography{bib/abbreviation-short,
    bib/abbreviation-empty,
    bib/all,
    bib/new_references}
}


\end{document}